\documentclass[conference,a4paper]{APSIPA2012}
\usepackage{multirow}
\usepackage{graphicx}
\usepackage{amsmath}
\usepackage[psamsfonts]{amssymb}
\usepackage{amsxtra}
\usepackage{subcaption}
\begin{document}
\title{A Deep Learning Approach to Drone Monitoring}

\author{%
\authorblockN{%
Yueru Chen, 
Pranav Aggarwal, 
Jongmoo Choi, and
C.-C. Jay Kuo
}
\authorblockA{%
University of Southern California, California, USA \\
E-mail: \{yueruche, pvaggarw, jongmooc\}@usc.edu, cckuo@sipi.usc.edu}
}

\maketitle
\thispagestyle{empty}

\begin{abstract}
A drone monitoring system that integrates deep-learning-based detection
and tracking modules is proposed in this work.  The biggest challenge in
adopting deep learning methods for drone detection is the limited amount
of training drone images. To address this issue, we develop a
model-based drone augmentation technique that automatically generates
drone images with a bounding box label on drone's location.  To track a
small flying drone, we utilize the residual information between
consecutive image frames. Finally, we present an integrated detection
and tracking system that outperforms the performance of each individual
module containing detection or tracking only. The experiments show that,
even being trained on synthetic data, the proposed system performs well
on real world drone images with complex background. The USC drone
detection and tracking dataset with user labeled bounding boxes is
available to the public. 
\end{abstract}

\section{Introduction}

There is a growing interest in the commercial and recreational use of
drones. This in turn imposes a threat to public safety.  The Federal
Aviation Administration (FAA) and NASA have reported numerous cases of
drones disturbing the airline flight operations, leading to near
collisions. It is therefore important to develop a robust drone
monitoring system that can identify and track illegal drones.  Drone
monitoring is however a difficult task because of diversified and
complex background in the real world environment and numerous drone
types in the market.  

Generally speaking, techniques for localizing drones can be categorized
into two types: acoustic and optical sensing techniques.  The acoustic
sensing approach achieves target localization and recognition by using a
miniature acoustic array system.  The optical sensing approach processes
images or videos to estimate the position and identity of a target
object. In this work, we employ the optical sensing approach by
leveraging the recent breakthrough in the computer vision field. 

The objective of video-based object detection and tracking is to detect
and track instances of a target object from image sequences. In earlier
days, this task was accomplished by extracting discriminant features
such as the scale-invariant feature transform (SIFT) \cite{sift} and the
histograms of oriented gradients (HOG) \cite{hog}.  The SIFT feature
vector is attractive since it is invariant to object's translation,
orientation and uniform scaling. Besides, it is not too sensitive to
projective distortions and illumination changes since one can transform an
image into a large collection of local feature vectors. The HOG feature
vector is obtained by computing normalized local histograms of image
gradient directions or edge orientations in a dense grid. It provides
another powerful feature set for object recognition. 

In 2012, Krizhevsky {\em et al.} \cite{alexnet} demonstrated the power
of the convolutional neural network (CNN) in the ImageNet grand
challenge, which is a large scale object classification task,
successfully. This work has inspired a lot of follow-up work on the
developments and applications of deep learning methods. A CNN consists
of multiple convolutional and fully-connected layers, where each layer
is followed by a non-linear activation function. These networks can be
trained end-to-end by back-propagation.  There are several variants in
CNNs such as the R-CNN \cite{rcnn}, SPPNet \cite{sppnet} and Faster-RCNN
\cite{fasterrcnn}.  Since these networks can generate highly
discriminant features, they outperform traditional object detection
techniques in object detection by a large margin. The Faster-RCNN
includes a Region Proposal Network (RPNs) to find object proposals, and
it can reach real time computation. 

The contributions of our work are summarized below.
\begin{itemize}
\item To the best of our knowledge, this is the first one to use the
deep learning technology for the challenging drone detection and
tracking problem. 
\item We propose to use a large number of synthetic drone images, which
are generated by conventional image processing and 3D rendering algorithms,
along with a few real 2D and 3D data to train the CNN. 
\item We propose to use the residue information from an image sequence
to train and test an CNN-based object tracker. It allows us to track a
small flying object in a cluttered environment. 
\item We propose an integrated drone monitoring system that consists of a
drone detector and a generic object tracker. The integrated system
outperforms the detection-only and the tracking-only sub-systems. 
\item We have validated the proposed system on several drone datasets.
\end{itemize}

The rest of this paper is organized as follows. The collected drone
datasets are introduced in Sec. \ref{sec:dataset}. The proposed drone
detection and tracking system is described in Sec.  \ref{sec:solution}.
Experimental results are presented in Sec.  \ref{sec:results}.
Concluding remarks are given in Sec.  \ref{sec:conclusion}. 

\section{Data Collection and Augmentation}\label{sec:dataset} 

\begin{figure}[ht]
\begin{center}
    \begin{subfigure}[t]{0.5\textwidth} 
        \centering
        \includegraphics[width=70mm]{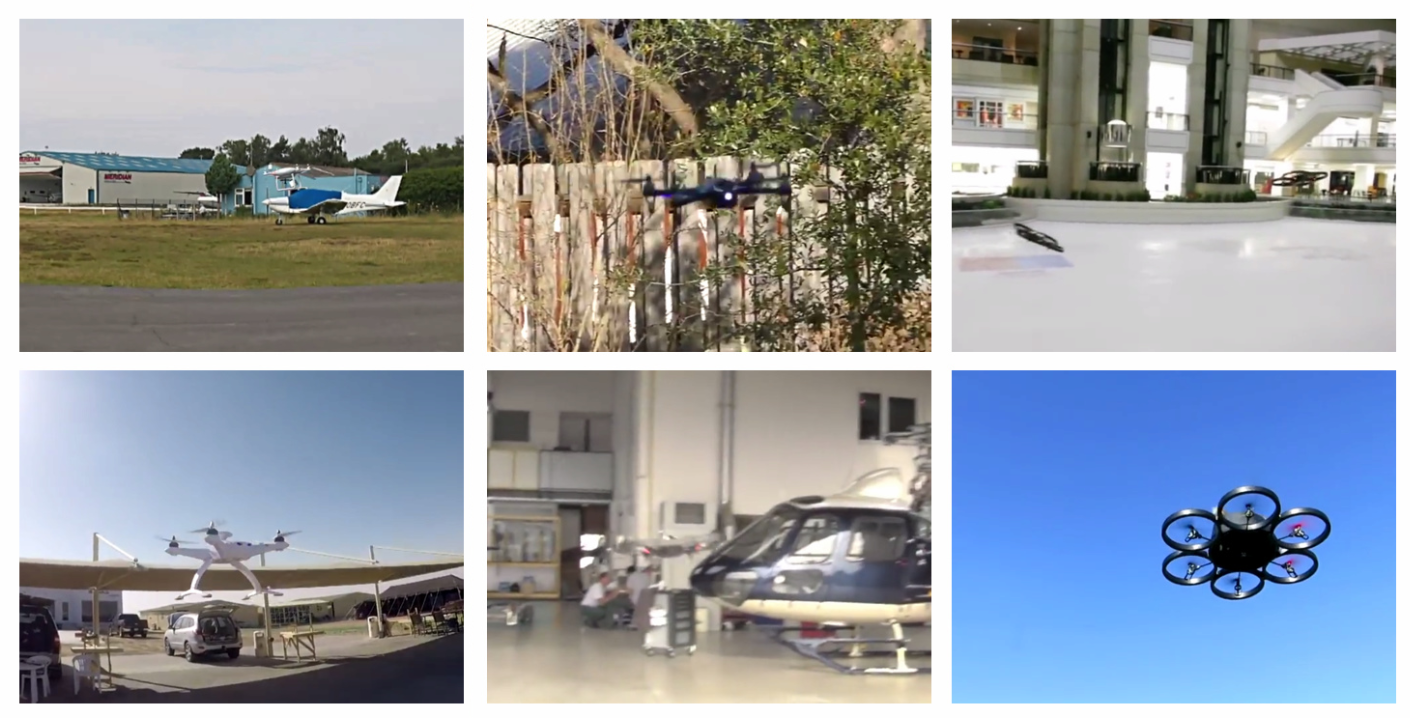}
        \caption{Public-Domain Drone Dataset} \label{fig:dataset2}
    \end{subfigure}
    \begin{subfigure}[t]{0.5\textwidth} 
        \centering
        \includegraphics[width=70mm]{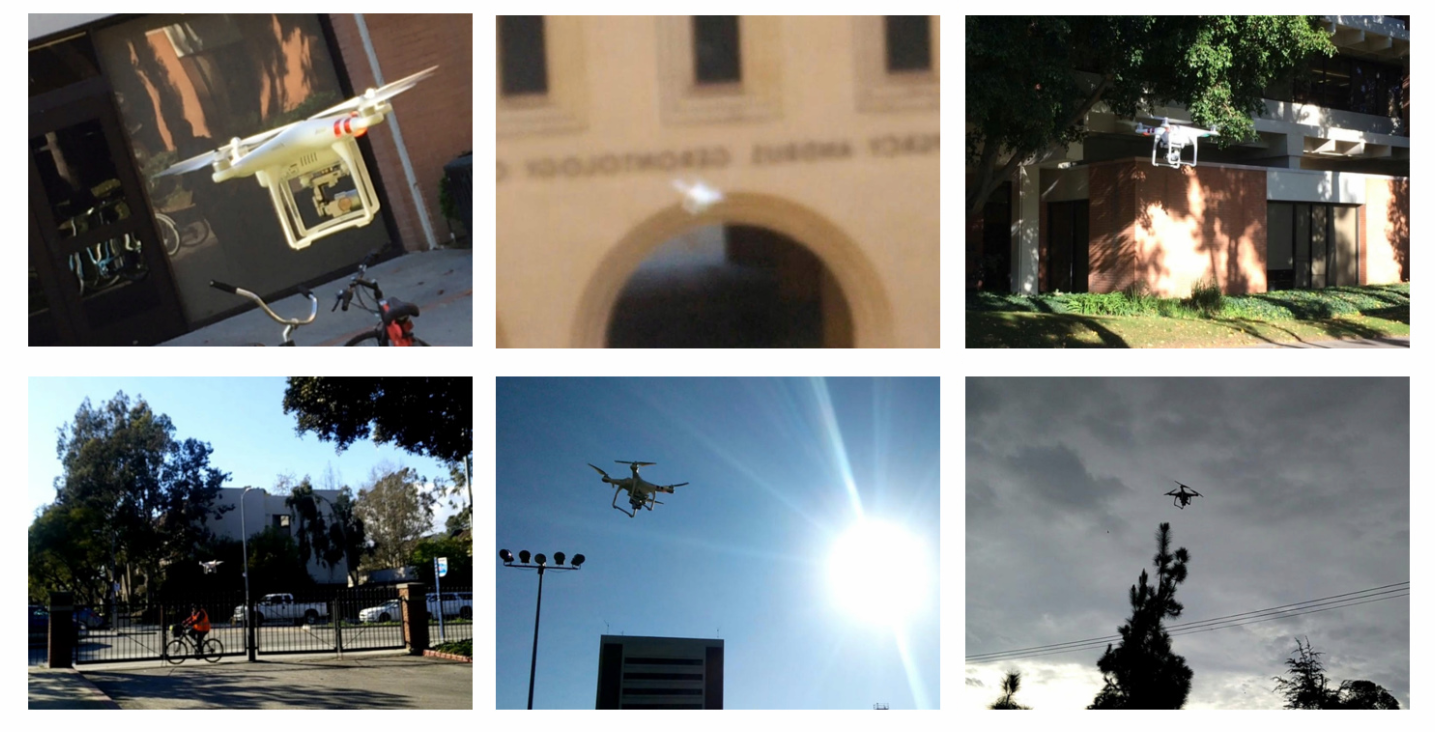}
        \caption{USC Drone Dataset} \label{fig:dataset1}
    \end{subfigure}
\end{center}
\caption{Sampled frames from two collected drone datasets.}\label{fig:dataset}
\end{figure}

\subsection{Data Collection}

The first step in developing the drone monitoring system is to collect
drone flying images and videos for the purpose of training and testing.
We collect two drone datasets as shown in Fig. \ref{fig:dataset}. They
are explained below.
\begin{itemize}
\item Public-Domain drone dataset. \\
It consists of 30 YouTube video sequences captured in an indoor or outdoor
environment with different drone models.  Some samples in this dataset are
shown in Fig.  \ref{fig:dataset2}. These video clips have a frame
resolution of 1280 x 720 and their duration is about one minute.  Some
video clips contain more than one drone. Furthermore, some shoots are
not continuous. 
\item USC drone dataset. \\
It contains 30 video clips shot at the USC campus. All of them were shot
with a single drone model. Several examples of the same drone in
different appearance are shown in Fig. \ref{fig:dataset1}. To shoot
these video clips, we consider a wide range of background scenes,
shooting camera angles, different drone shapes and weather conditions.
They are designed to capture drone's attributes in the real world such
as fast motion, extreme illumination, occlusion, etc.  The duration of
each video approximately one minute and the frame resolution is 1920 x
1080. The frame rate is 15 frames per second. 
\end{itemize}

We annotate each drone sequence with a tight bounding box around the
drone. The ground truth can be used in CNN training. It can also be used
to check the CNN performance when we apply it to the testing data.

\subsection{Data Augmentation}\label{sec:augmentation}

The preparation of a wide variety of training data is one of the main
challenges in the CNN-based solution.  For the drone monitoring task,
the number of static drone images is very limited and the labeling of
drone locations is a labor intensive job. The latter also suffers from
human errors. All of these factors impose an additional barrier in
developing a robust CNN-based drone monitoring system. To address this
difficulty, we develop a model-based data augmentation technique that
generates training images and annotates the drone location at each frame
automatically. 

\begin{figure}[t]
\begin{center}
\includegraphics[width=90mm]{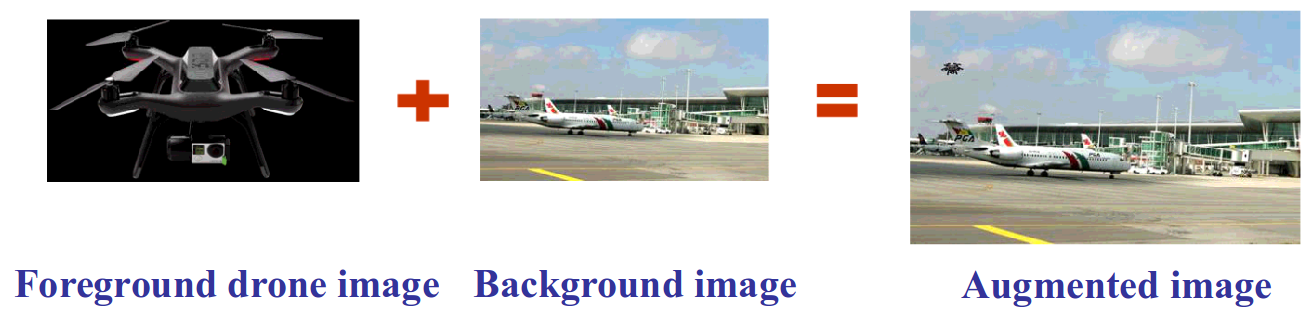}
\end{center}
\caption{Illustration of the data augmentation idea, where augmented
training images can be generated by merging foreground drone images and
background images.}\label{fig:augment}
\end{figure}

The basic idea is to cut foreground drone images and paste them on top
of background images as shown in Fig. \ref{fig:augment}. To accommodate
the background complexity, we select related classes such as aircrafts,
cars in the PASCAL VOC 2012 \cite{pascal}. As to the diversity of drone
models, we collect 2D drone images and 3D drone meshes of many drone
models. For the 3D drone meshes, we can render their corresponding
images by changing camera's view-distance, viewing-angle, lighting
conditions. As a result, we can generate many different drone images
flexibly.  Our goal is to generate a large number of augmented images to
simulate the complexity of background images and foreground drone models
in a real world environment.  Some examples of the augmented drone images of
various appearances are shown in Fig. \ref{fig:augment}. 

Specific drone augmentation techniques are described below.
\begin{itemize} 
\item Geometric transformations \\
We apply geometric transformations such as image translation, rotation
and scaling. We randomly select the angle of rotation from the range
(-30$^{\circ}$, 30$^{\circ}$). Furthermore, we conduct uniform scaling
on the original foreground drone images along the horizontal and the
vertical direction. Finally, we randomly select the drone location in
the background image. 
\item Illumination variation \\
To simulate drones in the shadows, we generate regular shadow maps by
using random lines and irregular shadow maps via Perlin noise
\cite{perlin}. In the extreme lighting environments, we observe that
drones tend to be in monochrome (i.e. the gray-scale) so that we 
change drone images to gray level ones. 
\item Image quality \\
This augmentation technique is used to simulate blurred drones caused
by camera's motion and out-of-focus. We use some blur filters (e.g.  the
Gaussian filter, the motion Blur filter) to create the blur effects on
foreground drone images. 
\end{itemize} 

\begin{figure}[ht]
\begin{center}
    \begin{subfigure}[t]{0.5\textwidth} 
        \centering
        \includegraphics[width=70mm]{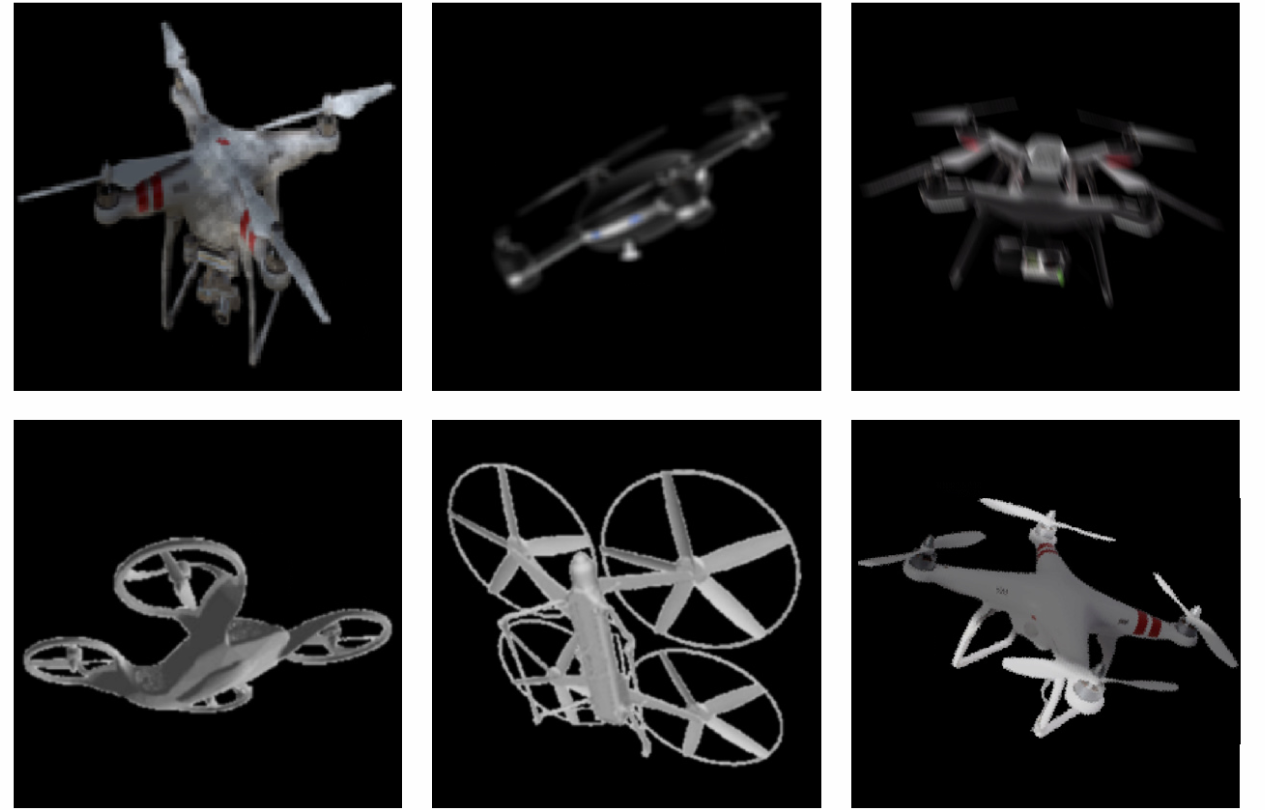}
        \caption{Augmented drone models} \label{fig:augdrone}
    \end{subfigure}
    \begin{subfigure}[t]{0.5\textwidth} 
        \centering
        \includegraphics[width=70mm]{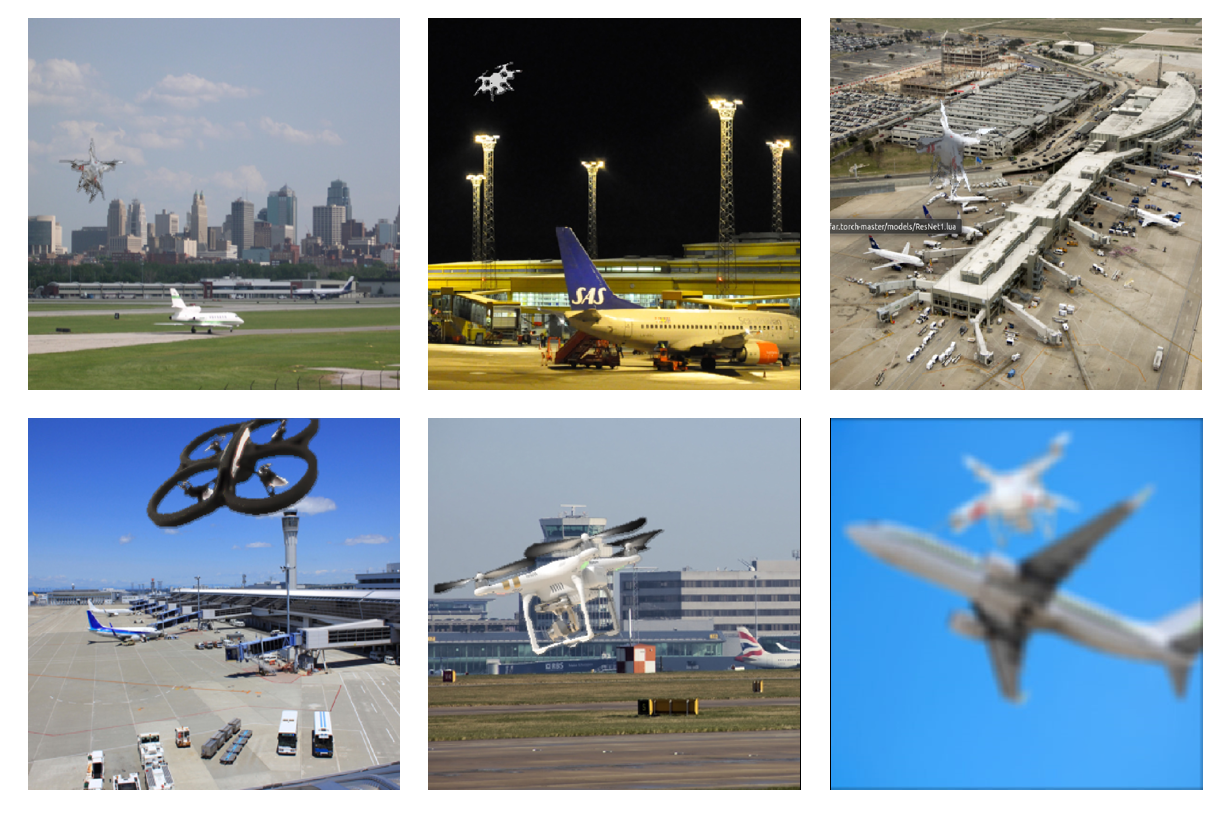}
        \caption{Synthetic training data} \label{fig:augimage}
    \end{subfigure}
\end{center}
\caption{Illustration of (a) augmented drone models and (b) synthesized
training images by incorporating various illumination conditions, image qualities,
and complex backgrounds.}\label{fig:augresult}
\end{figure} 

Several exemplary synthesized drone images are shown in Fig.
\ref{fig:augresult}, where augmented drone models are given in Fig.
\ref{fig:augdrone}.  We use the model-based augmentation technique to
acquire more training images with the ground-truth labels and show them
in Fig. \ref{fig:augimage}. 

\section{Drone Monitoring System}\label{sec:solution}

To realize the high performance, the system consists of two modules;
namely, the drone detection module and the drone tracking module. Both
of them are built with the deep learning technology. These two modules
complement each other, and they are used jointly to provide the accurate
drone locations for a given video input. 

\subsection{Drone Detection}\label{sec:detection}

The goal of drone detection is to detect and localize the drone in
static images. Our approach is built on the Faster-RCNN
\cite{fasterrcnn}, which is one of the state-of-the-art object detection
methods for real-time applications. The Faster-RCNN utilizes the deep
convolutional networks to efficiently classify object proposals. To
achieve real time detection, the Faster-RCNN replaces the usage of
external object proposals with the Region Proposal Networks (RPNs) that
share convolutional feature maps with the detection network. The RPN is
constructed on the top of convolutional layers. It consists of two
convolutional layers -- one that encodes conv feature maps for each
proposal to a lower-dimensional vector and the other that provides the
classification scores and regressed bounds. The Faster-RCNN achieves
nearly cost-free region proposals and it can be trained end-to-end by
back-propagation.  We use the Faster-RCNN to build the drone detector by
training it with synthetic drone images generated by the proposed data
augmentation technique as described in Sec. \ref{sec:augmentation}. 

\subsection{Drone Tracking}\label{sec:tracking}

The drone tracker attempts to locate the drone in the next frame based
on its location at the current frame.  It searches around the
neighborhood of the current drone's position.  This helps detect a drone
in a certain region instead of the entire frame. To achieve this
objective, we use the state-of-the-art object tracker called the
Multi-Domain Network (MDNet) \cite{mdnet}.  The MDNet is able to
separate the domain independent information from the domain specific
information in network training.  Besides, as compared with other
CNN-based trackers, the MDNet has fewer layers, which lowers the
complexity of an online testing procedure. 

\begin{figure}[]
\begin{center}
    \begin{subfigure}[t]{0.5\textwidth} 
        \centering
        \includegraphics[width=70mm]{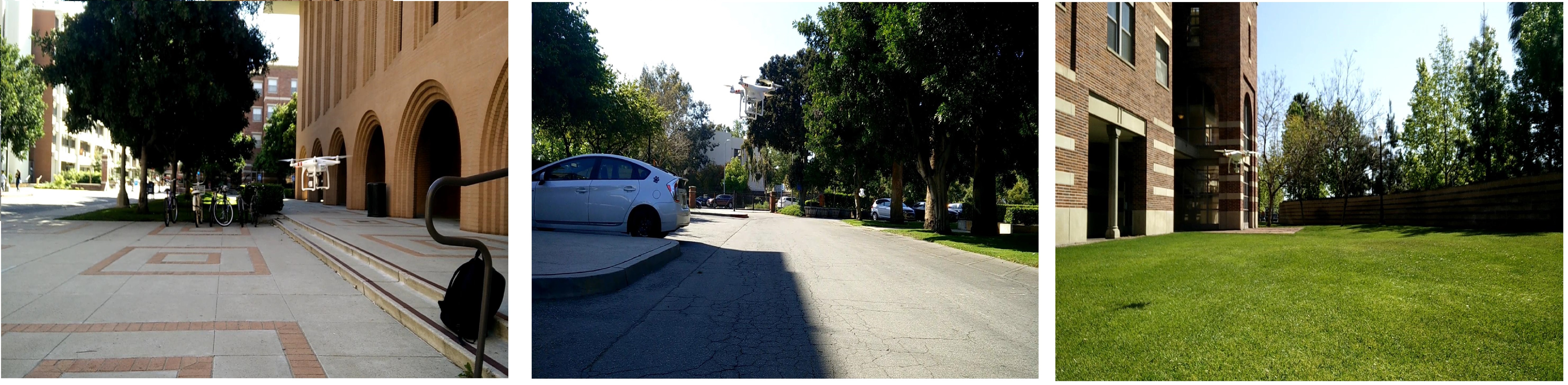}
        \caption{Raw input images} 
    \end{subfigure}
    \begin{subfigure}[t]{0.5\textwidth} 
        \centering
        \includegraphics[width=70mm]{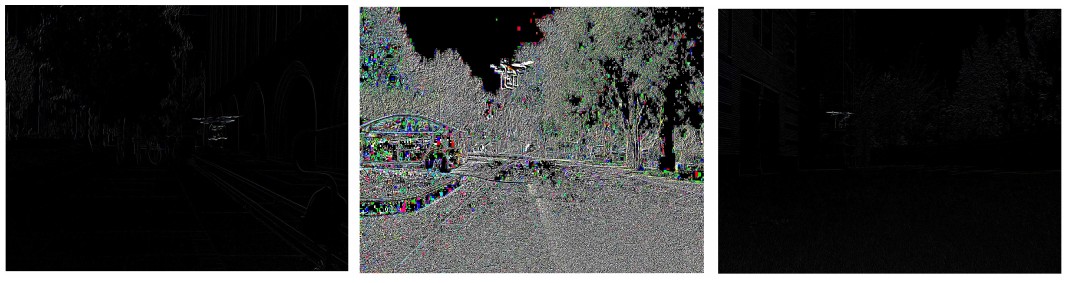}
        \caption{Corresponding residual images} 
    \end{subfigure}
\end{center}
\caption{Comparison of three raw input images and their corresponding
residual images.} \label{fig:residueimage}
\end{figure}

To improve the tracking performance furthermore, we propose a video
pre-processing step. That is, we subtract the current frame from the
previous frame and take the absolute values pixelwise to obtain the
residual image of the current frame.  Note that we do the same for the
R,G,B three channels of a color image frame to get a color residual
image.  Three color image frames and their corresponding color residual
images are shown in Fig.  \ref{fig:residueimage} for comparison.  If
there is a panning movement of the camera, we need to compensate the
global motion of the whole frame before the frame subtraction operation. 

Since there exists strong correlation between two consecutive images,
most background of raw images will cancel out and only the fast moving
object will remain in residual images. This is especially true when the
drone is at a distance from the camera and its size is relatively small.
The observed movement can be well approximated by a rigid body motion.
We feed the residual sequences to the MDNet for drone tracking after the
above pre-processing step. It does help the MDNet to track the drone
more accurately. Furthermore, if the tracker loses the drone for a short
while, there is still a good probability for the tracker to pick up the
drone in a faster rate. This is because the tracker does not get
distracted by other static objects that may have their shape and color
similar to a drone in residual images. Those objects do not appear in
residual images. 

\subsection{Integrated Detection and Tracking System}\label{sec:fusion}

\begin{figure}[ht]
\begin{center}
\includegraphics[width=70mm]{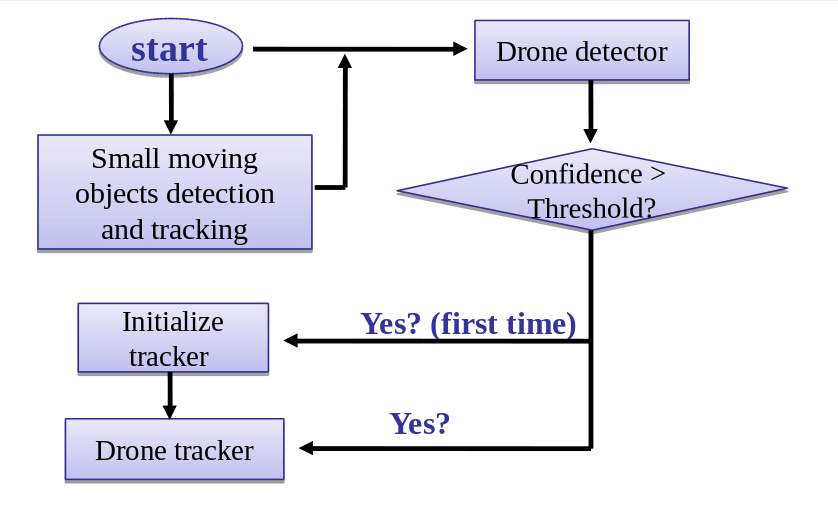}
\end{center}
\caption{A flow chart of the drone monitoring system.}\label{fig:overview}
\end{figure}

There are limitations in detection-only or tracking-only modules.  The
detection-only module does not exploit the temporal information, leading
to huge computational waste. The tracking-only module does not attempt
to recognize the drone object but follow a moving target only.  To build
a complete system, we need to integrate these two modules into one.  The
flow chart of the proposed drone monitoring system is shown in Fig.
\ref{fig:overview}. 

Generally speaking, the drone detector has two tasks -- finding the
drone and initializing the tracker. Typcially, the drone tracker is used
to track the detected drone after the initialization.  However, the
drone tracker can also play the role of a detector when an object is too
far away to be robustly detected as a drone due to its small size. Then,
we can use the tracker to track the object before detection based on the
residual images as the input. Once the object is near, we can use the
drone detector to confirm whether it is a drone or not. 

An illegal drone can be detected once it is within the field of view and
of a reasonable size. The detector will report the drone location to the
tracker as the start position. Then, the tracker starts to work.  During
the tracking process, the detector keeps providing the confidence score
of a drone at the tracked location as a reference to the tracker.  The
final updated location can be acquired by fusing the confidence scores
of the tracking and the detection modules as follows. 

For a candidate bounding box, we can compute the confidence scores of
this location via
\begin{eqnarray} \label{eqn:confidence}
    S'_d&=& 1 / ({1+e^{-\beta_1(S_d-\alpha_1)}}),\\
    S'_t&=& 1 / ({1+e^{-\beta_2(S_t-\alpha_2)}}),\\
    S' &=& \max(S'_d, S'_t),
\end{eqnarray}	
where $S_d$ and $S_t$ denote the confidence scores obtained by the
detector and the tracker, respectively, $S'_f$ is the confidence score
of this candidate location and parameters $\beta_1$, $\beta_2$,
$\alpha_1$, $\alpha_2$ are used to control the acceptance threshold. 

We compute the confidence score of a couple of bounding box candidates,
denoted by $BB_i$, $i \in C$, where $C$ denoted the set of candidate
indices. Then, we select the one with the highest score:
\begin{eqnarray}
i^* & = & \underset{i \in C}{\operatorname{argmax}}~S'_i, \\
S_f & = & \underset{i \in C}{\operatorname{max}}~S'_i, 
\end{eqnarray}	
where $BB_{i^*}$ is the finally selected bounding box and $S_f$ is its
confidence score. If $S_f = 0$, the system will report a message of
rejection. 

\section{Experimental Results}\label{sec:results}

\subsection{Drone Detection}

We test on both the real-world and the synthetic datasets. Each of them
contains 1000 images. The images in the real-world dataset are sampled from
videos in the USC Drone dataset. The images in the synthetic dataset are
generated using different foreground and background images in the
training dataset. The detector can take any size of images as the input.
These images are then re-scaled such that their shorter side has 600
pixels \cite{fasterrcnn}. 

\begin{figure}[t!]
\begin{center}
    \begin{subfigure}[t]{0.5\textwidth} 
        \centering
        \includegraphics[width=70mm]{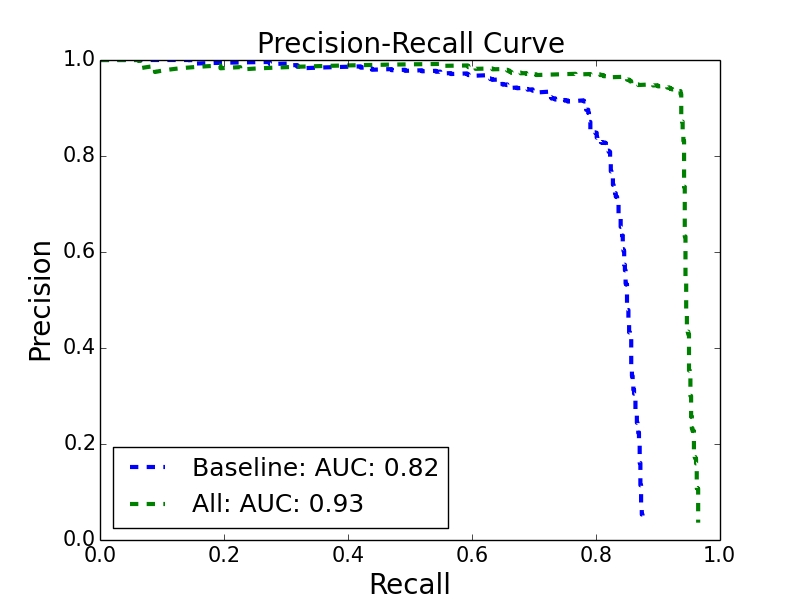}
        \caption{Synthetic Dataset} \label{fig:detectorRa}
    \end{subfigure}
    \begin{subfigure}[t]{0.5\textwidth} 
        \centering
        \includegraphics[width=70mm]{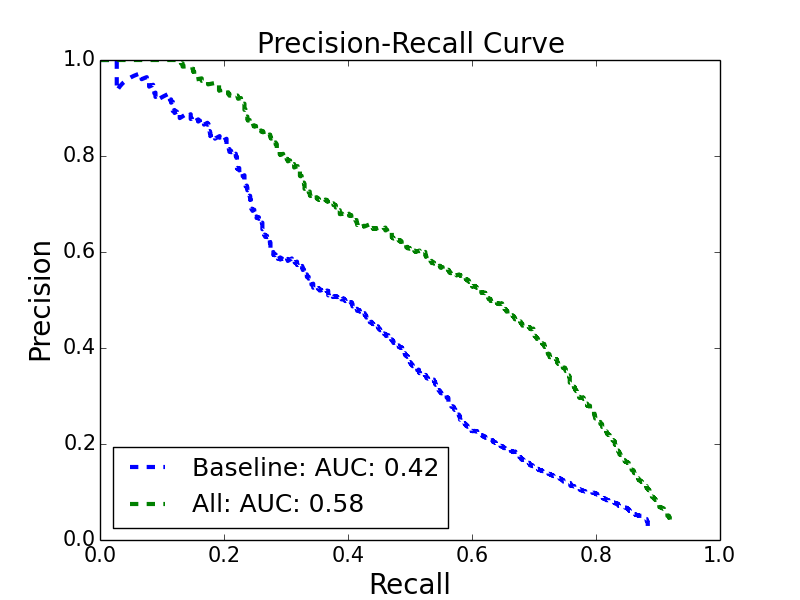}
        \caption{Real-World Dataset} \label{fig:detectorRb}
    \end{subfigure}
\end{center}
\caption{Comparison of the drone detection performance on (a) the
synthetic and (b) the real-world datasets, where the baseline method
refers to that uses geometric transformations to generate training data
only while the All method indicates that uses geometric transformations,
illumination conditions and image quality simulation for data
augmentation.} \label{fig:detectorR}
\end{figure}

To evaluate the drone detector, we compute the precision-recall curve.
Precision is the fraction of the total number of detections that are
true positive. Recall is the fraction of the total number of labeled
samples in positive class that are true positive. The area under the
precision-recall curve (AUC) \cite{auc} is also reported.  The
effectiveness of the proposed data augmentation technique is illustrated
in Fig. \ref{fig:detectorR}. In this figure, we compare the performance
of the baseline method that uses simple geometric transformations only
and that of the method that uses all mentioned data augmented
techniques, including geometric transformations, illumination conditions
and image quality simulation. Clearly, better detection performance can
be achieved by more augmented data.  We see around $11\%$ and $16\%$
improvements in the AUC measure on the real-world and the synthetic
datasets, respectively. 

\subsection{Drone Tracking}

The MDNet is adopted as the object tracker.  We take 3 video sequences
from the USC drone dataset as testing ones. They cover several
challenges, including scale variation, out-of-view, similar objects in
background, and fast motion.  Each video sequence has a duration of 30
to 40 seconds with 30 frames per second. Thus, each sequence contains
900 to 1200 frames. Since all video sequences in the USC drone dataset
have relatively slow camera motion, we can also evaluate the advantages
of feeding residual frames (instead of raw images) to the MDNet. 

The performance of the tracker is measured with the area-under-the-curve
(AUC) measure.  We first measure the intersection over union $(IoU)$ for all
frames in all video sequences as
\begin{equation}
IoU = \frac{Area~ of~ Overlap}{Area~ of~ Union}, 
\end{equation}	
where the ``Area of Overlap" is the common area covered by the predicted
and the ground truth bounding boxes and the ``Area of Union" is the
union of the predicted and the ground truth bounding boxes.  The IoU
value is computed at each frame. If it is higher than a threshold, the
success rate is set to 1; otherwise, 0.  Thus, the success rate value is
either 1 or 0 for a given frame.  Once we have the success rate values
for all frames in all video sequences for a particular threshold, we can
divide the total success rate by the total frame number. Then, we can
obtain a success rate curve as a function of the threshold. Finally, we
measure the area under the curve (AUC) which gives the desired
performance measure. 

\begin{figure}[t!]
\begin{center}
\includegraphics[width=90mm]{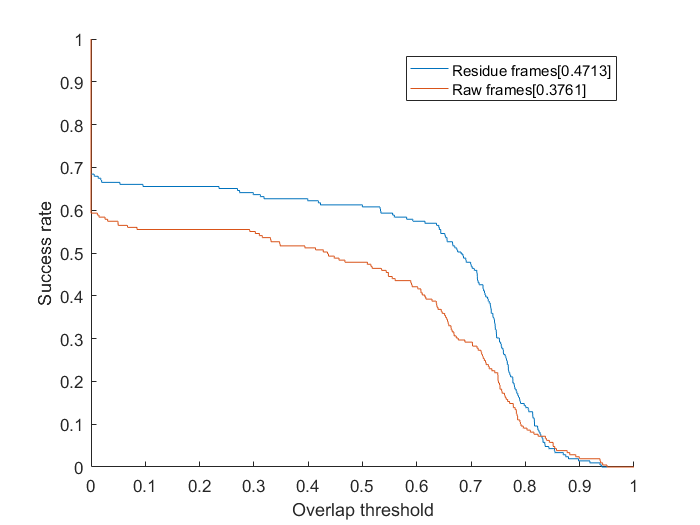}
\end{center}
\caption{Comparison of the MDNet tracking performance using the raw 
and the residual frames as the input.} \label{fig:trackR}
\end{figure}

We compare the success rate curves of the MDNet using the original
images and the residual images in Fig. \ref{fig:trackR}.  As compared to
the raw frames, the AUC value increases by around 10\% using the
residual frames as the input. It collaborates the intuition that
removing background from frames helps the tracker identify the drones
more accurately. Although residual frames help improve the performance
of the tracker for certain conditions, it still fails to give good
results in two scenarios: 1) movement with fast changing directions and
2) co-existence of many moving objects near the target drone.  To
overcome these challenges, we have the drone detector operating in
parallel with the drone tracker to get more robust results. 

\subsection{Fully Integrated System}

\begin{figure}[t!]
\begin{center}
       \includegraphics[width=90mm]{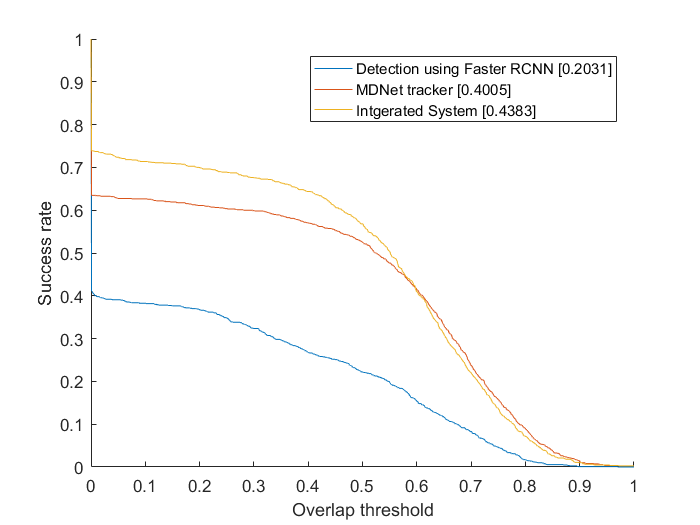}
\end{center}
\caption{Detection only (Faster RCNN) vs. tracking only (MDNet tracker)
vs. our integrated system: The performance increases when we fuse the
detection and tracking results.} \label{fig:systemR}
\end{figure}

The fully integrated system contains both the detection and the tracking
modules. We use the USC drone dataset to evaluate the performance of the
fully integrated system.  The performance comparison (in terms of the
AUC measure) of the fully integrated system, the conventional MDNet (the
tracker-only module) and the Faster-RCNN (the detector-only module) is
shown in Fig. \ref{fig:systemR}.  The fully integrated system
outperforms the other benchmarking methods by substantial margins. This
is because the fully integrated system can use detection as the means to
re-initialize its tracking bounding box when it loses the object. 

\section{Conclusion}\label{sec:conclusion}

A video-based drone monitoring system was proposed in this work.  The
system consisted of the drone detection module and the drone tracking
module. Both of them were designed based on deep learning networks.  We
developed a model-based data augmentation technique to enrich the
training data. We also exploited residue images as the input to the
drone tracking module. The fully integrated monitoring system takes
advantage of both modules to achieve high performance monitoring.
Extensive experiments were conducted to demonstrate the superior
performance of the proposed drone monitoring system. 

\section*{Acknowledgment}

This research is supported by a grant from the Pratt \& Whitney
Institute of Collaborative Engineering (PWICE). 



\begin{thebibliography}{10}
\bibitem{alexnet}
A.~Krizhevsky, I.~Sutskever, and G.~E. Hinton, ``Imagenet classification with
  deep convolutional neural networks,'' in \emph{ Advances in neural information
  processing systems}, pp.~1097--1105, 2012.
  
\bibitem{fasterrcnn}
S.~Ren, K.~He, R.~Girshick, and J.~Sun, ``Faster R-CNN: Towards real-time
  object detection with region proposal networks,'' in \emph{ Advances in neural
  information processing systems}, pp.~91--99, 2015.

\bibitem{rcnn}
R.~Girshick, J.~Donahue, T.~Darrell, and J.~Malik, ``Rich feature hierarchies
  for accurate object detection and semantic segmentation,'' in \emph{ Computer
  Vision and Pattern Recognition}, 2014.
  
\bibitem{sppnet}
K.~He, X.~Zhang, S.~Ren, and J.~Sun, ``Spatial pyramid pooling in deep
  convolutional networks for visual recognition,'' in \emph{ European Conference
  on Computer Vision}, pp.~346--361, Springer, 2014.

\bibitem{mdnet}
H.~Nam, B.~Han, \emph{Learning Multi-Domain Convolution Neural Networks for 
Visual Tracking}.In CVPR, 2016.  

\bibitem{auc}
J.~Huang and C.~X. Ling, ``Using AUC and accuracy in evaluating learning
  algorithms,'' \emph{ IEEE Transactions on knowledge and Data Engineering},
  vol.~17, no.~3, pp.~299--310, 2005.

\bibitem{hog}
N.~Dalal and B.~Triggs, ``Histograms of oriented gradients for human
  detection,'' in \emph{ Computer Vision and Pattern Recognition, 2005. CVPR
  2005. IEEE Computer Society Conference on}, vol.~1, pp.~886--893, IEEE, 2005.

\bibitem{sift}
D.~G. Lowe, ``Distinctive image features from scale-invariant keypoints,'' \emph{
  International journal of computer vision}, vol.~60, no.~2, pp.~91--110, 2004.

\bibitem{pascal}
M.~Everingham, L.~Van~Gool, C.~K.~I. Williams, J.~Winn, and A.~Zisserman, ``The
  {PASCAL} {V}isual {O}bject {C}lasses {C}hallenge 2012 {(VOC2012)}
  {R}esults.''
  http://www.pascal-network.org/challenges/VOC/voc2012/workshop/index.html.

\bibitem{perlin}
K.~Perlin, ``An image synthesizer,'' \emph{ ACM Siggraph Computer Graphics},
  vol.~19, no.~3, pp.~287--296, 1985.

\end{thebibliography}
\end{document}